\documentclass[letterpaper, 10 pt, conference]{ieeeconf}
\IEEEoverridecommandlockouts
\usepackage{cite}
\usepackage{amsmath,amssymb,mathrsfs,amsfonts}
\usepackage{bbm}
\usepackage{algorithmic}
\usepackage{graphicx}
\usepackage{subcaption}
\usepackage{color}
\usepackage[font=small]{caption}

\begin{document}

\title{\bf Locomotion and Obstacle Avoidance of a Worm-like Soft Robot}
    
\author{Sean Even, Yasemin Ozkan-Aydin,\textit{ Member, IEEE}
\thanks{*This work was not supported by any organization}
\thanks{$^{1}$All the authors are with the Department of Electrical Engineering, University of Notre Dame, Notre Dame, IN 46556 USA
        {\tt\small seven,yozkanay@nd.edu}}
}

\maketitle
\thispagestyle{empty}
\pagestyle{empty}

\begin{abstract}
This paper presents a soft earthworm robot that is capable of both efficient locomotion and obstacle avoidance. The robot is designed to replicate the unique locomotion mechanisms of earthworms, which enable them to move through narrow and complex environments with ease. The robot consists of multiple segments, each with its own set of actuators, that are connected through rigid plastic joints, allowing for increased adaptability and flexibility in navigating different environments. The robot utilizes proprioceptive sensing and control algorithms to detect and avoid obstacles in real-time while maintaining efficient locomotion. The robot uses a pneumatic actuation system to mimic the circumnutation behavior exhibited by plant roots in order to navigate through complex environments. The results demonstrate the capabilities of the robot for navigating through cluttered environments, making this development significant for various fields of robotics, including search and rescue, environmental monitoring, and medical procedures.

\end{abstract}

\section{Introduction}

Worm-like soft robots have garnered significant attention in the field of robotics due to their potential applications in search and rescue, environmental monitoring, and medical procedures \cite{application}. These robots are designed to replicate the unique locomotion mechanisms of earthworms, which allow them to move through narrow and complex environments with ease. In particular, the coordinated contraction and expansion of longitudinal and circular muscles via peristaltic motion enable earthworms to move forward and turn in different directions while also avoiding obstacles in their path \cite{trueman_1975}.

\begin{figure}[!t]
\centering
\includegraphics[width=9cm]{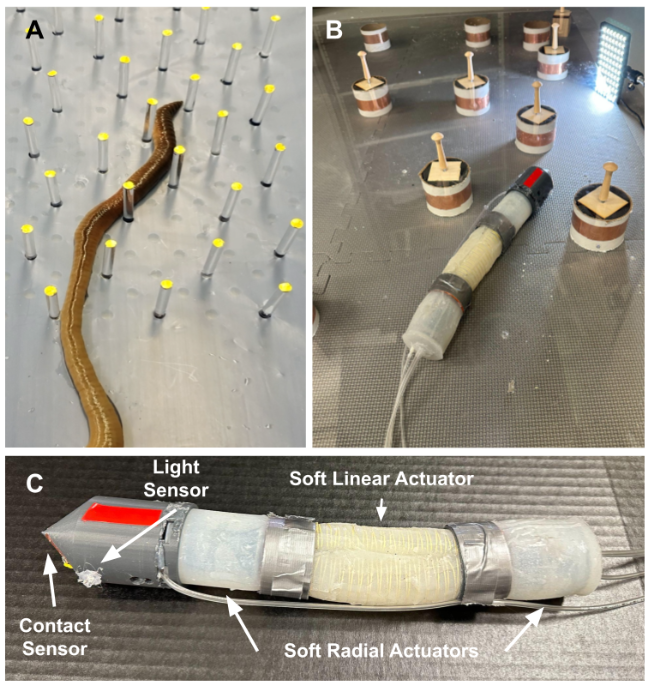}
\caption{\textbf{Illustration of obstacle navigation in biological earthworm and a soft earthworm robot. A.} Earthworm navigating through a challenging environment with rigid acrylic pegs regularly distributed on the surface. \textbf{B}. Our soft earthworm robot imitates this behavior. \textbf{C.} Feature description of our robot which includes one two-chambered linear soft actuator, two radial actuators, phototransistor light sensors, and copper tape contact sensors.  }
\label{fig:1}
\end{figure}

\begin{figure*}[!t]
\centering
\includegraphics[width=13cm]{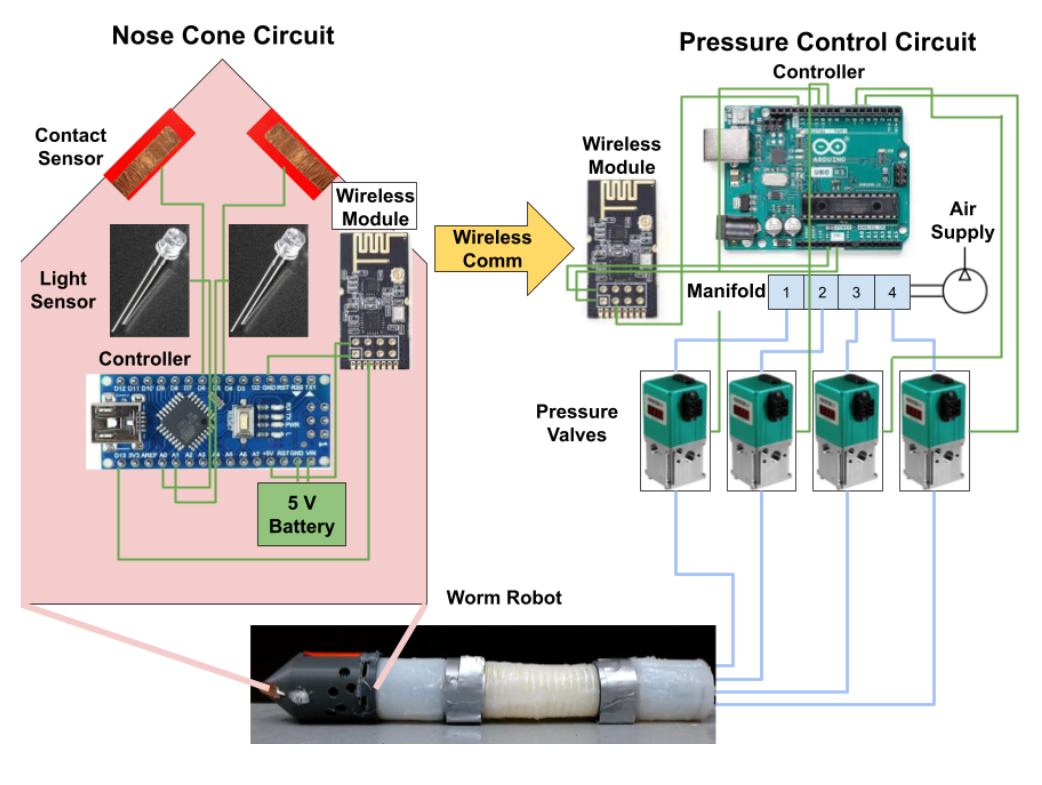}
\caption{\textbf{Control and sensing system of the robot}. The robot's nose cone circuit includes a microcontroller, phototransistor light sensors, and copper tape contact sensors. The sensory information is transmitted wirelessly to the pressure control circuit that includes an Arduino control board, four QB3 pneumatic regulators, an air supply, and a manifold to distribute the air. Note that electrical connections are noted in green and pneumatic connections are noted in blue.}
\label{fig:1}
\end{figure*}
Several soft robots inspired by natural locomotion have been developed to achieve different movements. Noorani et al. developed a crawling robot prototype based on the wave-like motion of caterpillar worms \cite{Noorani}. Niu et al. introduced a biomimetic worm-like robot, the MagWorm, with permanent magnetic patches for untethered crawling \cite{Niu}. Kalairaj et al. presented origami robots that are magnetically actuated and are capable of peristaltic, rolling, and turning motion \cite{kal}. Liu et al. designed a soft robot that combines Kirigami skin and radially expanding pneumatic actuators to mimic earthworm anchoring mechanisms \cite{Yas}. Xu et al. developed a mag-bot soft robot with thermoresponsive actuation and anisotropic friction, which can carry loads up to three times its weight \cite{Xu}. Du et al. proposed a crawling soft robot with a multi-movement mode, utilizing pneumatic actuators and auxetic cavity structures to achieve both straight-line and turning locomotion \cite{Du}. Das et al. developed a modular soft robot based on a peristaltic soft actuator inspired by earthworms, which can perform peristaltic locomotion in different media, providing insights into locomotion in unstructured subterranean environments \cite{Das2023}.

In this paper, we implement a mobile, worm-like soft robot that can  adapt its locomotion based on sensory inputs. Despite having no foreknowledge of the obstacles that block the robot's path to its target, it utilizes a combination of an oscillatory locomotion cycles and proprioceptive feedback to reach its target. This provides the framework for navigating through unknown terrain, such as an underground environment, in a robust and reliable fashion.

Several researchers have explored obstacle avoidance and path-planning techniques for soft robots, particularly those that employ peristaltic locomotion. Wang et al.  investigated the use of modified RRT algorithms and elliptical path generators, respectively, to optimize path planning for worm-like robots \cite{Wang1}, \cite{Wang2}, while Kandhari et al. presented a geometric approach for peristaltic turning analysis \cite{Kandhari}. Gough et al. proposed a path-planning method for morphing soft robots using 3D Voronoi diagrams and Dijkstra's algorithm \cite{Gough}. Verma et al. introduced a hierarchical planning approach for dynamic environments, with economic constraints\cite{Verma}. Finally, Zhang et al. presented an approach that utilizes the 3D deformability of the Yoshimura-origami structure for earthworm-like robots, enabling the robot to perform turning and rising motions and expand its reachable workspace \cite{Zhang}. The papers collectively offer insights and techniques for improving the obstacle avoidance and path-planning capabilities of soft robots.

 Despite all these studies, little has been done to imitate an earthworm's ability to navigate through an environment when the location of obstacles is unknown. While navigating through subterranean environments, the location of obstacles is seldom known, so complex path-planning algorithms are not practical. To gain an understanding of how to navigate around unseen obstacles, we draw inspiration from a plant root, another organism that navigates underground via growth. 
 Taylor et. al studied the obstacle navigation of plant root structure with minimal sensory input and how the introduction of oscillation or circumnutation in locomotion can minimize the chances of getting stuck on an obstacle \cite{roots} However, these methods have not been specifically applied to worm-like soft robots. We utilize proprioceptive sensing and control algorithms to enable our modular soft robot to detect and avoid obstacles in real-time while maintaining efficient locomotion.

\begin{figure}[!t]
\centering
\includegraphics[width=8.5cm]{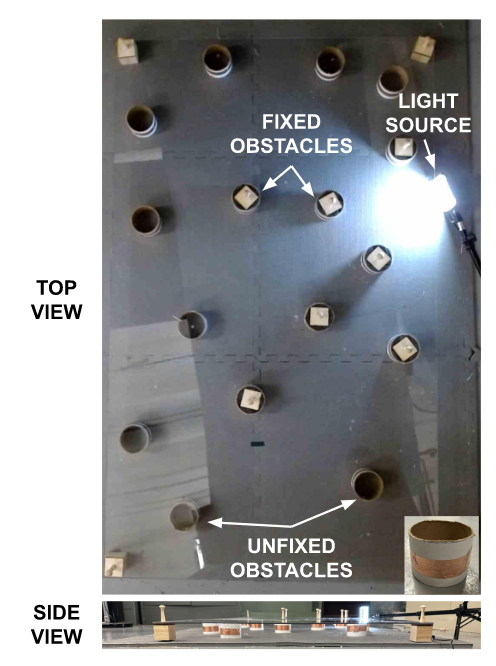}
\caption{ \textbf{The top and side view of the experimental test setup.} The setup consists of cylindrical pegs (d = 7.5 cm), six foam ground mats (60 cm x 60 cm) covered with an acrylic sheet (150 cm X 90 cm), and a light source. The pegs are covered The fixed obstacles are affixed to the structure to prevent the robot from colliding with them while traveling toward the light source. Conversely, the unfixed obstacles are positioned strategically to support the weight of the acrylic sheet. }
\label{fig:2}
\end{figure}
 Our modular robot consists of multiple segments, each with its own set o  actuators, that are connected through rigid plastic joints. This modular design allows for increased adaptability and flexibility in navigating different environments and allows for quick and easy reconfiguration of components, making it easier to customize and adapt the robot to different scenarios. We evaluate the performance of our robot in an obstacle environment and asses the ability of the robot to reach its target despite having a blocked path.

\section{Materials}
\subsection{Robot Fabrication}
\label{lab:robotFabrication}
In order to imitate the peristaltic gait of the earthworm robot, we designed a robot with three segments (Fig.\ref{fig:1}B-C). The front and back section has the ability to expand radially and the center section has the ability to expand longitudinally. 

The longitudinally-expansion actuators are cast with Dragon Skin 10 (Smooth-On Inc.) in 3-D printed molds (F170, Stratasys, Ltd.). The actuator consists of a pair of semi-circle 100 mm extensible chambers with a thickness of 2 mm and an outer diameter of 21 mm. After the first layer was cured, Kevlar thread was wrapped around the chamber in a double helix fashion at a pitch of 5.6 mm to focus the expansion of the actuator in the longitudinal direction. Then, another layer of Dragon Skin 10 (thickness 2mm) was added around all sides of the actuator to fix the location of the Kevlar threads and to seal holes in the first layer of casting. Finally, both ends were sealed with silicone ends caps which were each 3 mm thick.

The radial expansion actuators are similarly cast with Dragon Skin 20 (Smooth-On Inc.) in 3-D printed molds (F170, Stratasys, Ltd.). The outer layer of these actuators is a cylindrical 51 mm extensible chamber with an outer diameter of 46 mm and a thickness of 2 mm. Contained within each of these actuators is a 3-D printed backbone which is connected to the silicone outer layer with a thin layer of silicone. When pressure is added to the chamber, the plastic piece limits longitudinal expansion while it allows radial expansion. The outer layer of these actuators is 2 mm. The back actuator's backbone contains a cylindrical opening to allow the tubes from the center segment to feed through to the back of the robot. Each end of this actuator is sealed with a 5 mm disk of silicone. 

The segments of the robot are connected together using the press-fit 3-D printed rings and the 1/8 plastic tubes are fed through the robot. The nose cone was designed to have a conical shape to minimize the likelihood of catching on to obstacles. Located within this conical nose cone is the brain of the robot. This consists of an Arduino Nano Microcontroller, a 5V battery, two phototransistors that act as the robot's eyes, and two strips of copper tape which are used as contact sensors. The microcontroller can detect changes in capacitance in the copper wire. When the capacitance of the copper wire spikes, this signals that contact with the copper tape has been made. The phototransistors and copper tape are placed on opposite sides of the nose cone. Also included within this circuit is an nRF24L01 transceiver module information which allows the data from the sensors to be transmitted to the Pressure Control board where it can be interpreted to modulate pressure. 

\begin{figure}[!t]
\centering
\includegraphics[width=9cm]{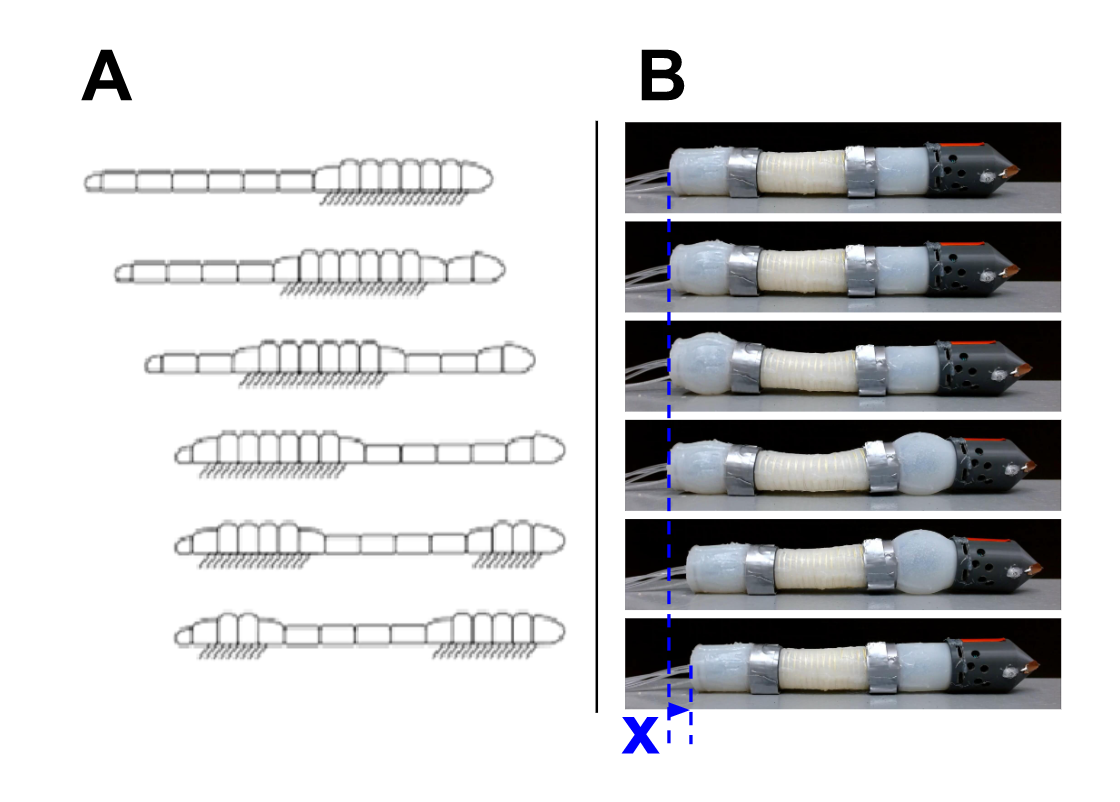}
\caption{\textbf{Peristaltic Gait of biological and robotic earthworm.} \textbf{A.} Earthworm moves from left to right with a peristaltic gait \cite{Perastaltic}, \textbf{B.} the locomotion cycle of an earthworm-like robot. The value of X is the net displacement due to one cycle of locomotion.}
\label{fig:3}
\end{figure}

\subsection{Test Setup}
\label{sec:testSetup}
The robot is actuated pneumatically by a Pneumatic Control Setup consisting of four QB3 pneumatic regulators, an Arduino Uno control board, and another nRF24L01 transceiver module which receives sensor information from the nosecone. The QB3 is a closed-loop pressure regulator made up of a mechanical regulator mounted to two solenoid valves, an internal pressure transducer, and electronic controls. By turning on the solenoid valves, which pressurize the mechanical regulator's pilot, the pressure is controlled. Both valves regulate the exhaust and the inlet respectively.

We tested the robot's ability to navigate a confined environment between a foam mat and a thin sheet of acrylic plastic (Fig. \ref{fig:2}) The acrylic sheet is 152 by 91 cm in dimension.
The acrylic sheet simulates the confined space where worms navigate through underground environments and helps with the anchoring of segments of the radial actuators during locomotion.  We also introduced circular obstacles randomly throughout the environment. These obstacles are made out of cardboard tubes and also wrapped in cardboard tape to give the obstacles capacitance. Finally, a light source is added to the environment. The goal of the robot is to navigate from its starting location to be as close to the light as possible. The robot must navigate through a field of obstacles in order to accomplish this task.

\section{Methods}
In this segment, we will describe the actuation methods and decision structure of the robot makes as it navigates through the testing environment given in Sec.\ref{sec:testSetup}.

\subsection{Locomotion}
Earthworms move through underground environments using a process called retrograde peristalsis (Fig.\ref{fig:3}) where the segments of the worm's body are expanded and extended in a coordinated fashion to achieve locomotion\cite{Perastaltic}.

\begin{figure}[!t]
\centering
\includegraphics[width=0.5\textwidth]{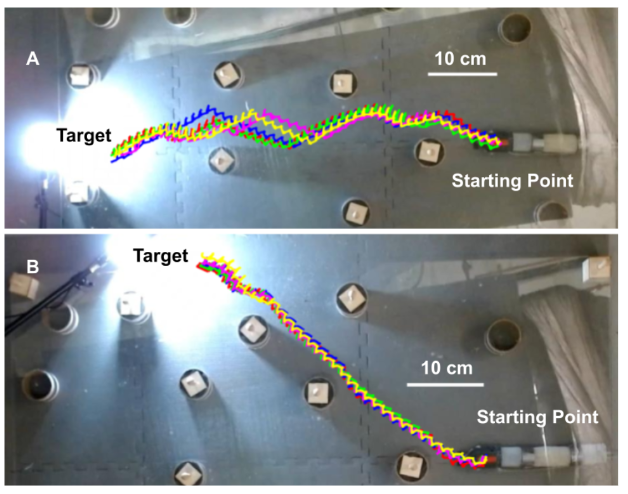}
\caption{ \textbf{Locomotion of the robot towards the light source while navigating obstacles. A.}The trajectories of the nose from five separate runs to reach the light source on the far side of the testing environment  \textbf{B.}Similarly, trajectories of five runs to reach a target at the top of a testing environment.}
\label{fig:5}
\end{figure}
As described in Sec.\ref{lab:robotFabrication}, our robot is outfitted with segments capable of expansion and contraction. In order to imitate the gait pattern of earthworms, the robot is actuated in the following cycle.

In forward locomotion, first, pressure is applied to the back radial actuator in order to anchor the robot. Next, the center segment extends forward while remaining anchored in the back. Once, the center finishes extending, the front radial actuator expands while the back radial actuator contracts. After the front actuator is in place, the center segment contracts while it is held in place by the front anchor. At this point, the front anchor contracts, and the cycle can start over again. The displacement caused by one locomotion cycle referred to as X in Fig.\ref{fig:3}, is most greatly impacted by the amount of expansion in the radial and longitudinal actuators. Of these two parameters, however, it is most impactful to maximize the expansion achievable by the center actuator.

We also want the robot to have the ability to turn left and right. This is possible because there are two semi-circular chambers within the center segment. When we want the robot to move straight, both chambers of the center segment are actuated at the same time. However, when we want the robot to turn, we just actuated one of these chambers at a time. If we want the robot to turn right, we turn on the left actuator and if we want the robot to turn left, we turn on the right actuator. This is because when we apply pressure to a chamber, it extends while the other side does not. The effect of this is bending toward the unactuated side, which is why we must actuate it in the other direction. Other than that, the locomotion cycle is identical. 

\subsection{Sensing and Control}

The nose cone of the worm robot contains contact sensors as well as phototransistors. The circuit in the nose cone is constantly sending information back to the Pneumatic Control Board. The sensor information is transmitted back to the Pneumatic Control board at a frequency of 5 Hz. The highest priority information to transmit is if the nose cone has made contact with an obstacle. When this happens, the nose cone circuit tells when it happens and on which side it occurs. When there is no contact with the copper tape, the light sensor information is transmitted. Light value readings are constantly taken from the phototransistors on either side of the robot and compared against one another. The photo transmitter with a higher light value wins out and  this is communicated to the Pneumatic Control Board respectively.

Previous studies have used external vision methods to give more information about the robot, but we would like to move towards a completely mobile robot. Thus, we rely only on proprioceptive information that can be obtained by onboard sensors. 

As the pneumatic control board receives the information from the sensors, it makes decisions about what should happen in the next cycle. If it receives light sensor information. one-directional locomotion cycle occurs steering the worm robot toward the light and then one forward locomotion cycle occurs propelling the robot forward. One result of this strategy is that there is a slight oscillation in the tip of the robot as it moves. As discussed in the introduction, this behavior in roots is called circumnutation and it significantly reduces the chances of getting stuck on an obstacle \cite{roots}. 

When contact with an obstacle is detected, the robot's goal is to steer toward  the obstacle. This behavior allows the robot to push off the obstacle and continue on its way toward the light source.  Thus when contact is detected on either side of the robot, two-directional cycles occur away from the direction in which contact occurs. Then, one forward locomotion cycle occurs. At this point, if the robot is still in contact with the obstacle, the process repeats. If not, it resumes looking for the light and moving toward it.

\section{Results}
We hypothesized that in order to navigate toward the light the robot would need to steer toward it. However, empirical observations during the testing phase provided evidence to the contrary. Upon turning towards the light, at the end of its cycle, it would come to rest skewed slightly in the other direction. This unexpected outcome arose due to the interaction between the front section of the robot and the acrylic sheet during the gait cycle. Due to the elevated level of friction, it was observed that the front portion of the robot would catch on the acrylic sheet, subsequently propelling the robot in an opposite direction than originally intended. Although this is not an expected outcome, the behavior is predictable. So, we merely needed to invert the response in order to reach our goal.

 It was determined that the most effective response was achieved through the application of precise pressure levels and duration to the radial and center actuators. However, it is worth noting that due to fabrication errors, the pressure levels of the two front and back radial actuators were not set to equal values. Specifically, the optimal response was obtained when a pressure of 28 kPa was applied to the back radial actuators for a duration of one second, a pressure of 55 kPa was applied to each of the center actuators for a duration of two seconds, a pressure of 42 kPa was applied to the back radial actuators for a duration of one second in each actuation cycle. 

In experiments, we placed a light source in different positions around the setup that would act as a target destination for the worm robot. Two representative sets of successful experiments were given in Fig. \ref{fig:5}. In the first set (Fig. \ref{fig:5}A), the robot is attempting to reach a light source on the back wall of the testing environment. In all five attempts, the robot successfully makes it to the target despite colliding with multiple obstacles with a progress of $8.22\pm 0.16 mm$ per cycle. Similarly, in the second set (Fig. \ref{fig:5}B) a target is on the right top of the testing environment. In this set of tests, on average the robot approaches the target by $10.08 \pm 0.27$ mm per cycle. In this test, the robot has more of a direct path to the target, while previously moving vertically was merely to avoid obstacles.

Note that although all the tests in Fig.\ref{fig:5} show the robot successfully converging to the target, this was not consistently observed. The robot's limited turning radius, coupled with the presence of obstacles, occasionally resulted in deviations from the intended trajectory toward the target. This occurred in roughly a third of obstacle orientations attempted where the robot was inadvertently pushed off the converging course due to collisions with obstacles.

The worm robot, despite encountering numerous obstacles in all the tests conducted, managed to avoid getting stuck on an obstacle. The body undulation motion caused by the alternation between a directional locomotion cycle and a forward locomotion cycle contributed to its success. There were instances where the robot required several cycles to overcome a particular obstacle, yet invariably it prevailed.

\section{Conclusion and Future Work}

In conclusion, our research successfully developed a soft robotic worm capable of efficient locomotion and obstacle avoidance in a confined environments. Through extensive testing, we identified the optimal actuation parameters and demonstrated that our robot was able to successfully converge towards a light source despite colliding with multiple obstacles.

While our current design relies on a pneumatic tether, our future work aims to eliminate this constraint by incorporating miniature pumps into the robot itself. This would enable our robot to be entirely mobile and expand its range of applications.

Our results have significant implications for various fields of robotics, especially for navigating through complex environments where traditional robots may struggle. By utilizing a modular design and proprioceptive sensing, our robot has the potential to explore subterranean environments in a more efficient and effective manner.


\section{Acknowledgement}
We would like to acknowledge Asraf Siddique for his help with the silicone mold design of the radial and linear actuators.
\bibliographystyle{IEEEtran}

\end{document}